\pdfoutput=1

\documentclass[11pt]{article}

\usepackage[]{acl}

\usepackage{times}
\usepackage{latexsym}

\usepackage[T1]{fontenc}

\usepackage[utf8]{inputenc}

\usepackage{microtype}

\usepackage{booktabs}
\usepackage{graphicx}
\usepackage{enumitem}
\usepackage{multirow}
\usepackage{makecell}
\usepackage{xspace}
\usepackage{todonotes}
\usepackage{amsmath}
\usepackage{amsfonts}
\usepackage{tabularx}
\usepackage{pifont}
\usepackage{enumitem}

\newcommand\OurDataset{\emph{CrowdChecked}\xspace}
\newcommand\CLEFDataset{\emph{CheckThat~'21}\xspace}

\newcommand{\cmark}{\ding{51}}%
\newcommand{\xmark}{\ding{55}}%

\definecolor{lincolngreen}{rgb}{0.11, 0.35, 0.02}
\definecolor{brickred}{rgb}{0.8, 0.25, 0.33}

\newcommand{\correctmark}{{\color{lincolngreen}{\cmark}}}
\newcommand{\incorrectmark}{\color{brickred}{\xmark}}

\title{\emph{\OurDataset}: Detecting Previously Fact-Checked Claims in Social Media}

\author{Momchil Hardalov$^1$ \quad Anton Chernyavskiy$^2$ \\
\textbf{Ivan Koychev$^1$ \quad Dmitry Ilvovsky$^2$ \quad Preslav Nakov$^3$} \\
  $^1$Sofia University ``St. Kliment Ohridski'', Bulgaria \\
  $^2$HSE University, Russia \\
  $^3$Mohamed bin Zayed University of Artificial Intelligence, UAE \\
  {\tt \{hardalov, koychev\}@fmi.uni-sofia.bg}\\
  {\tt \{acherniavskii, dilvovsky\}@hse.ru} \\
  {\tt  preslav.nakov@mbzuai.ac.ae}
}

\begin{document}
\maketitle
\begin{abstract}
While there has been substantial progress in developing systems to automate fact-checking, they still lack credibility in the eyes of the users. Thus, an interesting approach has emerged: to perform automatic fact-checking by verifying whether an input claim has been previously fact-checked by professional fact-checkers and to return back an article that explains their decision. This is a sensible approach as people trust manual fact-checking, and as many claims are repeated multiple times.
Yet, a major issue when building such systems is the small number of known {tweet--verifying article} pairs available for training. Here, we aim to bridge this gap by making use of crowd fact-checking, i.e., mining claims in social media for which users have responded with a link to a fact-checking article. In particular, we mine a large-scale collection of 330,000 tweets paired with a corresponding fact-checking article. We further propose an end-to-end framework to learn from this noisy data based on modified self-adaptive training, in a distant supervision scenario. Our experiments on the CLEF'21 CheckThat! test set show improvements over the state of the art by two points absolute. Our code and datasets are available at \url{https://github.com/mhardalov/crowdchecked-claims}
\end{abstract}
\section{Introduction}
\label{sec:introduction}

The massive spread of disinformation online, especially in social media, was counter-acted by major efforts to limit the impact of false information not only by journalists and fact-checking organizations but also by governments, private companies, researchers, and ordinary Internet users. This includes building systems for automatic fact-checking ~\citep{zubiaga-etal-2016-rumor-spread,derczynski-etal-2017-rumoureval,Survey:2021:AI:Fact-Checkers,EMNLP2022:PASTA,guo-etal-2022-survey,hardalov-etal-2022-survey}, 
fake news~\citep{ferreira-vlachos-2016-emergent,FANG:ACM}, and fake news website detection~\cite{baly-etal-2020-written,stefanov-etal-2020-predicting,EMNLP2022:GREENER}.

\begin{figure}[t]
    \centering
    \includegraphics[width=\columnwidth]{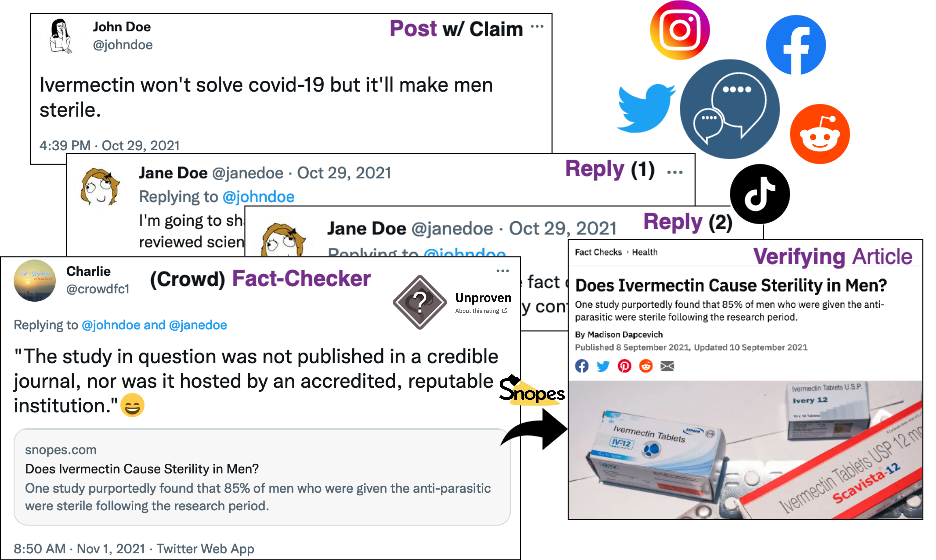}
    \caption{
    Crowd fact-checking thread on Twitter. 
    {The first tweet ({\bf Post w/ claim}) makes the claim that \emph{Ivermectin causes sterility in men}, which then receives \textbf{replies}. A \textbf{(crowd) fact-checker} replies with a link to a \textbf{verifying article} from a fact-checking website. We pair the \emph{article} with the \emph{tweet that made this claim} (the first post {\correctmark}), as it is irrelevant ({\incorrectmark}) to the other \emph{replies}.
    }
    }
    \label{fig:motivation}
\end{figure}

Unfortunately, fully automatic systems still lack credibility, and thus it was proposed to focus on detecting previously fact-checked claims instead:
\emph{Given a user comment, detect whether the claim it makes was previously fact-checked with respect to a collection of verified claims and their corresponding articles} (see Table~\ref{tab:task_examples}). This task is an integral part of an end-to-end fact-checking pipeline~\cite{10.14778/3137765.3137815}, and also an important task on its own right as people often repeat the same claim~\cite{barron2020overview,vo-lee-2020-facts,DBLP:conf/clef/ShaarHMHBAMEN21}. Research on this problem is limited by data scarceness, with datasets typically having about a 1,000 tweet--verifying article pairs \cite{barron2020overview,shaar-etal-2020-known,DBLP:conf/clef/ShaarHMHBAMEN21}, with the notable exception of \cite{vo-lee-2020-facts}, which contains 19K claims about images matched against 3K fact-checking articles.

We propose to bridge this gap using crowd fact-checking
to create a large collection of tweet--verifying article pairs, which we then label (if the pair is correctly matched) automatically using distant supervision. An example is shown in Figure~\ref{fig:motivation}.

Our contributions are as follows:

\begin{itemize}
    \item we mine a large-scale collection of 330,000 tweets paired with fact-checking articles;
    \item we propose two distant supervision strategies to label the \emph{\OurDataset} dataset;
    \item we propose a novel method to learn from this data using modified self-adaptive training;
    \item we demonstrate sizable improvements over the state of the art on a standard test set.
\end{itemize}

\section{Our Dataset: \OurDataset}
\label{sec:dataset}

\subsection{Dataset Collection}
\label{subsec:datacreate}

We use Snopes as our target fact-checking website,
due to its popularity among both Internet users and
researchers~\cite{popat2016credibility,hanselowski-etal-2019-snopes,augenstein-etal-2019-multifc,tchechmedjiev2019claimskg}. We further use Twitter as the source for collecting user messages, which could contain claims and fact-checks of these claims.

Our data collection setup is similar to the one in \cite{10.1145/3331184.3331248}.
First, we form a query to select tweets that contain a link to a fact-check from Snopes (\emph{url:snopes.com/fact-check/}), which is either a reply or a quote tweet, and not a retweet.
An example result from the query is shown in Figure~\ref{fig:motivation}, where the tweet \emph{from the crowd fact-checker} contains a link to a fact-checking article. We then assess its relevance to the claim (if any) made in the first tweet (the root of the conversation) and the last reply in order to obtain tweet--verified article pairs. We analyze in more detail the conversational structure of these threads in Section~\ref{sec:appx:convstruct}.

\begin{table}[t!]
    \centering
    
    \resizebox{.5\textwidth}{!}{%
    \small
    \begin{tabular}{p{0.01\textwidth}ll}
        \toprule
        \multicolumn{3}{c}{\makecell[l]{\parbox{\columnwidth}{{\bf User Post w/ Claim}: Sen. Mitch McConnell: ``As recently as October, now-President Biden said you can't legislate by executive action unless you are a dictator. Well, in one week, he signed more than 30 unilateral actions.'' [URL] — Forbes (@Forbes) January 28, 2021}}} \\
        \\
        \multicolumn{3}{c}{\bf Verified Claims and their Corresponding Articles} \\
        \\
        (1) & \makecell[l]{\parbox{0.82\columnwidth}{When he was still a candidate for the presidency in \\October 2020, U.S. President Joe Biden said, “You can't legislate by executive order unless you're a dictator.”  { \url{http://snopes.com/fact-check/biden-executive-order-dictator/}}}} & \correctmark \\ 
        \\
        (2) & \makecell[l]{\parbox{0.82\columnwidth}{U.S. Sen. Mitch McConnell said he would not participate in 2020 election debates that include female moderators. { \url{http://snopes.com/fact-check/mitch-mcconnell-debate-female/}}}} & \incorrectmark \\ 
        \bottomrule
    \end{tabular}
    }
    \caption{{Illustrative examples for the task of detecting previously fact-checked claims. The \textbf{post contains a claim} (related to \emph{legislation and dictatorship}), the \textbf{Verified Claims} are part of a search collection of previous fact-checks. In row (\emph{1}), the fact-check is a correct match for the claim made in the tweet (\correctmark),  whereas in (\emph{2}), the claim still discusses \emph{Sen. Mitch McConnell}, but it is a different claim ({\incorrectmark}), and thus this is an incorrect pair.}}
    \label{tab:task_examples}
    \vspace{-0.2cm}
\end{table}

We collected all tweets matching our query from October 2017 till October 2021, obtaining a total of 482,736 unique hits. We further collected 148,503 reply tweets and 204,250 conversation (root) tweets.\footnote{The sum of the unique replies and of the conversation tweets is not equal to the total number of fact-checking tweets, as more than one tweet might reply to the same comment. 
}
Finally, we filter out malformed pairs, i.e.,~tweets linking to themselves, empty tweets, non-English ones, such with no resolved URLs in the Twitter object (\emph{`entities'}), with broken links to the fact-checking website, and all tweets in the \emph{\CLEFDataset} dataset. 
We ended up with 332,660 unique tweet--article pairs (shown in first row in Table~\ref{tab:datasplits}), 316,564 unique tweets, and 10,340 fact-checking articles from Snopes they point to.

More detail about the process of collecting fact-checking articles as well as detailed statistics are given in Appendix~\ref{sec:appx:factcheckcollect} and on Figure~\ref{fig:snopes_dist}.

\subsection{Tweet Collection} (Conversation Structure)
\label{sec:appx:convstruct}
It is important to note that the \emph{`fact-checking'} tweet can be part of a multiple-turn conversational thread, therefore taking the post that it replies to (previous turn), does not always express a claim which the current tweet targets. In order to better understand this, we performed manual analysis of some conversational threads. Conversational threads in Twitter are organized as shown Figure~\ref{fig:motivation}: the root is the first comment, then there can be a long discussion, followed by a fact-checking comment (i.e., the one with a link to a fact-checking article on Snopes). In our analysis, we identify four patterns: (\emph{i})~the current tweet verifies a claim in the tweet it replies to, (\emph{ii})~the tweet verifies the root of the conversation, (\emph{iii})~the tweet does not verify any claim in the chain (a common scenario), and (\emph{iv})~the fact-check targets a claim that was not expressed in the root or in the closest tweet (this was in very few cases). This analysis suggests that for the task of detecting previously fact-checked claims, it is sufficient to collect the triplet of the fact-checking tweet, the root of the conversation (\emph{conversation}), and the tweet that the target tweet is replying to (\emph{reply}).

\begin{table}[t]
    \centering
    \setlength{\tabcolsep}{4pt}
    \resizebox{.5\textwidth}{!}{%
    \begin{tabular}{l|rrrrrrr}
    \toprule
    \bf Dataset      &   \bf{Tweets}$^{\ddagger}$         & \multicolumn{3}{c}{\bf{Words}} & \bf{Vocab} \\
     &    $|$Unique$|$       & Mean & 50\% & Max & $|$Unique$|$ \\
    \midrule
    \bf{\OurDataset (Ours)} &    316,564 &      12.2 &     11 &  60 &  114,727 \\
    \CLEFDataset &      1,399 &     17.5 &     16 &  62 &    9,007 \\
    \bottomrule
    \end{tabular}
    }
    \caption{Statistics about our dataset vs. \CLEFDataset. $^{\ddagger}$The number of unique tweets is lower than the total number of tweet--article pairs, as an input tweet could be fact-checked by multiple articles.}
    \label{tab:clef_compare_vocab}
\end{table}

\subsection{Comparison to Existing Datasets}

We compare our dataset to a closely related dataset from the CLEF-2021 \CLEFDataset on Detecting Previously Fact-Checked Claims in Tweets \citep{DBLP:conf/clef/ShaarHMHBAMEN21}, to which we will refer as \emph{\CLEFDataset} in the rest of the paper. There exist other related datasets that are  smaller~\cite{barron2020overview}, come from a different domain~\cite{DBLP:conf/clef/ShaarHMHBAMEN21}, are not in English~\cite{elsayed2019overview}, or are multi-modal~\citep{vo-lee-2020-facts}.

Table~\ref{tab:clef_compare_vocab} compares our \emph{\OurDataset} to \emph{\CLEFDataset} in terms of number of examples, length of the tweets, and vocabulary size. Before calculating these statistics, we lowercased the text and we removed all URLs, Twitter handlers, English stop words, and punctuation. We can see in Table~\ref{tab:clef_compare_vocab} that \emph{\OurDataset} contains two orders of magnitude more examples, slightly shorter tweets (but the maximum length stays approximately the same, which can be explained by the word limit of Twitter), and has a vocabulary size that is an order of magnitude larger.
Note, however, that many examples in \emph{\OurDataset} are incorrect matches (see Section~\ref{subsec:datacreate}), and thus we use distant supervision to label them (see Section~\ref{sec:datalabelling}), with the resulting dataset sizes of matching pairs shown in Table~\ref{tab:datasplits}. Here, we want to emphasize that there is absolutely no overlap at all between \OurDataset and \CLEFDataset in terms of tweets/claims.

In terms of topics, the claims in both our dataset and \emph{\CLEFDataset} are quite diverse, including fact-checks for a broad set of topics related, but not limited to politics (e.g.,~the Capitol Hill riots, US elections), pop culture (e.g.,~famous performers and actors such as Drake and Leonardo di Caprio), brands (e.g.,~McDonald's and Disney), and COVID-19, among many others. Illustrative examples of the claim/topic diversity can be found in Tables~\ref{tab:task_examples} and \ref{tab:annotation_examples} (in the Appendix). Moreover, the collection of Snopes articles contains almost 14K different fact-checks on an even wider range of topics, which further diversifies the set of tweet--article pairs.

Finally, we compare the set of Snopes fact-checking articles referenced by the crowd fact-checkers to the ones included in the \CLEFDataset competition. We can see that the tweets in \OurDataset refer to less articles (namely 10,340), compared to \CLEFDataset, which consists of 13,835 articles. A total of 8,898 articles are present in both datasets. Since the \CLEFDataset is collected earlier, it includes less articles from recent years compared to \OurDataset, and peaks at 2016/2017. Nevertheless, for \CLEFDataset, the number of Snopes articles included in a claim--article pair is far less compared to our dataset (even after filtering out unrelated pairs), as it is capped at the number of tweets included in that dataset (which is 1.4K).

\begin{figure}[t]
    \centering
    \includegraphics[width=\columnwidth]{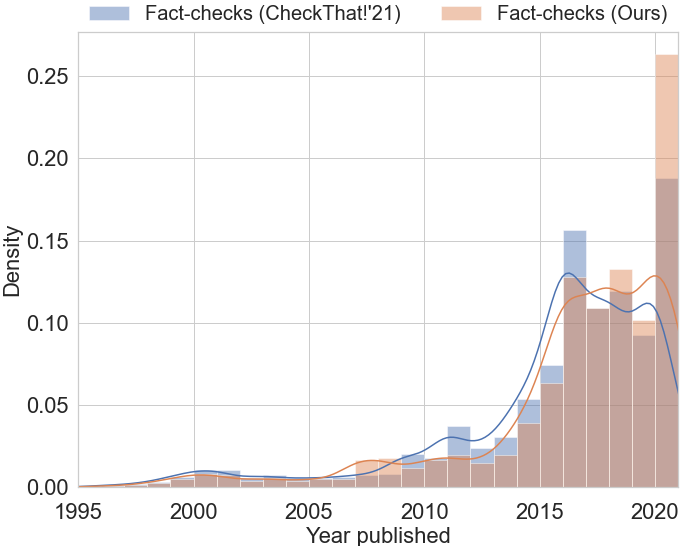}
    \caption{Histogram of the year of publication of the Snopes articles included in \emph{\OurDataset} (our dataset) vs. those in \emph{\CLEFDataset}.}
    \label{fig:snopes_dist}
\end{figure}

More detail about the process of collecting the fact-checking articles is given in Appendix~\ref{sec:appx:factcheckcollect}.

\subsection{Data Labeling (Distant Supervision)}
\label{sec:datalabelling}

To label our examples,
we experiment with two distant supervision approaches: (\emph{i})~based on the Jaccard similarity between the tweet and the target fact-checking article,
and (\emph{ii})~based on the predictions of a model trained on \CLEFDataset.

\paragraph{Jaccard Similarity} In this approach, we first pre-process the texts by converting them to lowercase, removing all URLs and replacing all numbers with a single zero. Then, we tokenize them using NLTK's \emph{Twitter tokenizer}~\cite{loper-2002-nltk}, 
and we strip all handles and user mentions. Finally, we filter out all stop words
and punctuation (including quotes and special symbols) and we stem
all tokens using the Porter stemmer~\cite{porter1980algorithm}.

In order to obtain a numerical score for each tweet--article pair, we calculate the \emph{Jaccard similarity} (jac) between the normalized tweet text and each of the \emph{title} and the \emph{subtitle} from the Snopes article (i.e.,~the intersection over the union of the unique tokens). Both fields present a summary of the fact-checked claim, and thus should include more compressed information. Finally, we average these two similarity values to obtain a more robust score. Statistics are shown in Table~\ref{tab:wo_data_stats}.

\begin{table}[t]
    \centering
    \setlength{\tabcolsep}{3pt}
    \resizebox{.5\textwidth}{!}{%
    \begin{tabular}{lrrr}
    \toprule
    \bf Range &    \bf Examples  & \bf {Correct} Pairs &    \bf {Correct} Pairs \\
    \small{(Jaccard)} & (\%)& Reply (\%)  & Conv. (\%)  \\
    \midrule
    {[}0.0;0.1) &	 62.57 &   5.88    &   0.00 \\
    {[}0.1;0.2) &	 18.98 &  36.36    &  14.29 \\
    {[}0.2;0.3) &	 10.21 &  46.67    &  50.00 \\
    {[}0.3;0.4) &	  4.17 &  76.47    &  78.57 \\
    {[}0.4;0.5) &	  2.33 &  92.86    &  92.86 \\
    {[}0.5;0.6) &	  1.08 &  94.12    &  94.12 \\
    {[}0.6;0.7) &	  0.43 &  80.00    &  80.00 \\
    {[}0.7;0.8) &	  0.11 &  92.31    &  92.31 \\
    {[}0.8;0.9) &     0.05 &  91.67    &  92.86 \\
    {[}0.9;1.0{]} &	  0.02 &  100.00   & 100.00 \\
    \bottomrule
    \end{tabular}
    }
    \caption{Proportion of examples in different bins based on average Jaccard similarity between the tweet and the title/subtitle. 
    Manual annotations of the \textit{correct pairs.}}
    \label{tab:wo_data_stats}
\end{table}

\paragraph{Semi-Supervision} Here, we train a SentenceBERT~\cite{reimers-gurevych-2019-sentence} model, as described in Section~\ref{sec:method:sbert}, using the manually annotated data from \CLEFDataset. The model shows strong performance on the testing set of \CLEFDataset (see Table~\ref{tab:sbert:distsup}), and thus we expect it to have good precision at detecting matching fact-checked pairs. In particular, we calculate the \emph{cosine similarity} between the embeddings of the fact-checked tweet and the fields from the Snopes article. Statistics about the scores are shown in Table~\ref{tab:cos_data_stats}.

\begin{table}[t]
    \centering
    \setlength{\tabcolsep}{3pt}
    \begin{tabular}{lrr}
    \toprule
           \bf Range &   \bf Examples  & \bf {Correct} Pairs \\
           \small{(Cosine)} & (\%) & (\%)\\
    \midrule
     {[}-0.4;0.1) &       37.83 &         0.00 \\
      {[}0.1;0.2) &       16.50 &         6.67 \\
      {[}0.2;0.3) &       12.28 &         41.46 \\
      {[}0.3;0.4) &       10.12 &         36.36 \\
      {[}0.4;0.5) &        8.58 &         63.16 \\
      {[}0.5;0.6) &        6.69 &         70.00 \\
      {[}0.6;0.7) &        4.47 &         84.21 \\
      {[}0.7;0.8) &        2.48 &         96.15 \\
      {[}0.8;0.9) &        0.97 &         93.10 \\
    {[}0.9;1.0{]} &        0.08 &        100.00 \\
    \bottomrule
    \end{tabular}
    \caption{Proportion of examples in different bins based on cosine similarity using Sentence-BERT trained on \emph{\CLEFDataset}. Manual annotations of the \textit{correct pairs.}}
    \label{tab:cos_data_stats}
\end{table}

\subsection{Feasibility Evaluation} 
\label{sec:feasability}

To evaluate the feasibility of the obtained labels, we performed manual annotation, aiming to estimate the number of \emph{correct pairs} (i.e.,~tweet--article pairs, where the article fact-checks the claim in the tweet). Our prior observations of the data suggested that unbiased sampling from the pool of tweets was not suitable, as it would include mostly pairs that have very few overlapping words, which is 
often an indicator that the texts are not related. Thus, we sample the candidates for annotation based on their Jaccard similarity.

We divided the range of possible values [0;1] into 10 equally sized bins and we sampled 15 examples from each bin, resulting into 150 conversation--reply--tweet triples. Afterwards, the appropriateness of each reply-article and conversation-article pair is annotated by three annotators independently.
The annotators had a \emph{good level} of inter-annotator agreement: 0.75 
in terms of Fleiss Kappa~\cite{fleiss1971measuring} (see Appendix~\ref{sec:appx:annotations}).

Tables~\ref{tab:wo_data_stats} and~\ref{tab:cos_data_stats} show the resulting estimates of \emph{correct pairs} for both Jaccard and cosine-based labeling. In the case of Jaccard, we can see that the expected number of correct examples is very high (over 90\%) in the range of \emph{[0.4--1.0]}, and then it drastically decreases, going to almost zero when the similarity is less than 0.1. Similarly, for the cosine score, we can see high number of matches in the top 4 bins (\emph{[0.6--1.0]}), albeit the number of matches remains relatively high in the following interval of \emph{[0.2--0.6)} between 36\% and 63\%, and again gets close to zero for the lower-score bins. We analyze the distribution of the Jaccard scores in \CLEFDataset in more detail in Appendix~\ref{sec:appx:dataanalysis}.

\section{Method}
\label{sec:method}
\label{sec:method:sbert}

\paragraph{General Scheme}

As a base for our models, we use Sentence-BERT (SBERT). It uses a Siamese network trained with a Transformer~\cite{nips2017_7181:transformer} encoder 
to obtain sentence-level embeddings. We keep the base architecture proposed by \citet{reimers-gurevych-2019-sentence}, but we use additional features, training tricks, and losses described in the next sections.

Our input is a pair of a tweet and a fact-checking article, which we encode as follows:

\begin{itemize}%
    \item Tweet: 
    \textrm{[CLS] \emph{Tweet Text} [SEP]}
    \item {Verifying} article:
    \textrm{[CLS] \emph{Title} [SEP] \emph{Subtitle} [SEP] \emph{Verified Claim} [SEP]}
\end{itemize}

We train the models using the Multiple Negatives Ranking (MNR) loss~\cite{Henderson2017EfficientNL} (see Eq.~\ref{eq:1}), instead of the standard cross-entropy (CE) loss, as the datasets contain only positive (i.e.,~matching) pairs. Moreover, we propose a new variant of the MNR loss that accounts for the noise in the dataset, as described in detail in Section~\ref{sec:trainingwithnoise}.

\paragraph{Enriched Scheme}

In the enriched scheme of the model, we adopt the pipeline proposed in the best-performing system from the \CLEFDataset competition~\cite{DBLP:conf/clef/ChernyavskiyIN21}. Their method consists of independent components for assessing lexical (TF.IDF-based) and semantic (SBERT-based) similarities. The SBERT models use the same architecture and input format as described in the \emph{General Scheme} above. However, \citet{DBLP:conf/clef/ChernyavskiyIN21} use an ensemble of models, i.e.,~instead of calculating a single similarity between the tweet and the joint title/subtitle/verified claim, the similarities between the tweet and the claim, the joint title/claim, and the three together are obtained from three models, one using TF.IDF and one using SBERT, for each combination. These similarities are combined via a re-ranking model (see Section~\ref{sec:method:reranking}). In our experiments, the TF.IDF and the model ensembles are included only in the models with re-ranking.

\paragraph{Shuffling and Temperature} 
Additionally, we adopt a temperature parameter ($\tau$) in the MNR loss. We also make it trainable in order to stabilize the training process as suggested in \citep{chernyavskiy-etal-2022-batch}.
This forces the loss to focus on the most complex and most important examples in the batch. Moreover, this effect is amplified after each epoch by an additional data shuffling that composes batches from several groups of the most similar examples. This shuffling, in turn, increases the temperature significance. The nearest neighbors forming the groups are found using the model predictions.
More detail about the training and the models themselves
can be found in \citep{DBLP:conf/clef/ChernyavskiyIN21}.

\subsection{Training with Noisy Data}
\label{sec:trainingwithnoise}

\paragraph{Self-Adaptive Training} To account for possible noise in the distantly supervised data, we 
propose a new method based on self-adaptive training~\cite{huang2020self},
which was introduced for classification tasks and the CE loss; however, it needs to be modified in order be used with the MNR loss.
We iteratively refurbish the labels $y$ using the predictions of the current model starting after an epoch of choice, which is a hyper-parameter:

\begin{center}
   $ y^{r} \leftarrow \alpha \cdot y^r + (1 - \alpha) \cdot \hat{y}$, 
\end{center}

\noindent where 
$y^r$ is the current refurbished label ($y_r = y$ initially), $\hat{y}$ is the model prediction, and $\alpha$ is a momentum hyper-parameter (we set $\alpha$ to 0.9). 

Since the MNR loss operates with positive pairs only (it does not operate with labels), to implement this approach, we had to modify the loss function. Let $\{c_i, v_i\}_{i=1,\ldots,m}$ be the batch of input pairs, where $m$ is the batch size, $C,V \in \mathbb{R}^{m\times{h}}$ are the matrices of embeddings for the tweets and for the fact-checking articles ($h$ is the embeddings' hidden size), and $C, V$ are normalized to the unit hyper-sphere (we use cosine similarity), then:
\begin{equation} \label{eq:1}
\begin{split}
\mathcal{L} & =
-\frac{1}{m}\sum_{i=1}^{m}{y^{r}}_i\Big (\frac{c_i^T v_i}{\tau} - \log\sum_{j=1}^{m}\exp(\frac{c_i^T v_j}{\tau})\Big)
\end{split}
\end{equation}
If we set $y^{r}_i = 1$, then Eq.~\ref{eq:1} resembles the MNR loss definition. The parameter $\tau$ is the temperature, discussed in Section~\ref{sec:method:sbert} \emph{Shuffling and Temperature}. 

\paragraph{Weighting} 
In the self-adaptive training approach, \citet{huang2020self} introduce weights $w_i = \max_{j \in \{1,..,L\}}{t_{i,j}}$, where $t_i$ is the corrected one-hot encoded target vector in a classification task with $L$ classes. The goal is to ensure that noisy labels will have a lower influence on the training process compared to correct labels.
Instead of a classification task with one-hot target vectors $t_{i,j}$, here we have real targets $y^{r}_i$. Therefore, we take these probabilities as weights: $w_i = y^{r}_i$.
After applying both modifications with the addition of labels and weights, the impact of each training example is proportional to the square of the corrected label, i.e.,~in Eq.~\ref{eq:1} $y^{r}_i$ is now squared.

\subsection{Re-ranking} 
\label{sec:method:reranking}

Re-ranking has shown major improvements for detecting previously fact-checked claims~\cite{shaar-etal-2020-known,DBLP:conf/clef/ShaarHMHBAMEN21,DBLP:conf/clef/MihaylovaBCHHN21,DBLP:conf/clef/ChernyavskiyIN21}, and we include it as part of our model. 

In particular, we adopt the re-ranking procedure from \cite{DBLP:conf/clef/ChernyavskiyIN21}, which uses LambdaMART~\cite{Wu2009AdaptingBF} for re-ranking. The inputs are the reciprocal ranks (position in the ranked list of claims) and the predicted relevance scores (two factors) based on the scores of the TF.IDF and S-BERT models (two models), between the tweet and the claim, claim+title, and claim+title+subtitle (three combinations), for a total of twelve features in the ensemble and four in the single model.

\section{Experiments}
\label{sec:experiments}

In this section, we describe our experimental setup, baselines, and experimental results. The training procedure and the hyper-parameters are described in more detail in Appendix~\ref{sec:appx:hyper:ft}.

\subsection{Experimental Setup}

\paragraph{Datasets}
Table~\ref{tab:datasplits} shows statistics about the data split sizes for \OurDataset and \CLEFDataset. We use these splits in our experiments, albeit sometimes mixed together.

The first group (\OurDataset) is the data splits obtained using distant supervision. As the positive pairs are annotated with distant supervision and not by humans, we include them as part of the training set. Each shown split is obtained using a different similarity measure (Jaccard or Cosine) or threshold.
From the total number of 332K collected tweet--{article} pairs in \OurDataset, we ended up with subsets of sizes between 3.5K and 49K examples. 

The second group describes the \CLEFDataset dataset. We preserve the original training, development, and testing splits.
In each of our experiments, we validate and test on the corresponding subsets from the \CLEFDataset, while the training set can be a mix with \emph{\OurDataset}.

\paragraph{Evaluation Measures} We adopt the ranking measures used in the \CLEFDataset competition. In particular, we calculate the Mean Reciprocal Rank (MRR), Mean Average Precision (MAP@K), and Precision@K for $K \in \{1, 3, 5, 10\}$. We optimize our models for MAP@5, as was in the CLEF-2021 CheckThat! lab subtask 2A.

\begin{table}[t]
    \centering
    \setlength{\tabcolsep}{3pt}
    \resizebox{.48\textwidth}{!}{%
    \begin{tabular}{ll|rr}
    \toprule
        \bf Dataset & \bf Data Split & \bf Threshold &  \makecell[tr]{\bf Tweet-Article \\ \bf Pairs} \\
    \midrule
        \multirow{8}{*}{\makecell[cl]{\bf{\OurDataset}\\ (Our Dataset)}} & Train & - &    332,660 \\ \cmidrule{2-4}
        {} & \multirow{3}{*}{\makecell[cl]{Train \\\emph{Jaccard}}} & 0.30 &     27,387 \\
        {} & {} & 0.40 &     12,555 \\
        {} & {} & 0.50 &      4,953 \\ \cmidrule{2-4}
        {} & \multirow{4}{*}{\makecell[cl]{Train \\\emph{Cosine}}} & 0.50 &     48,845 \\
        {} & {} & 0.60 &     26,588 \\
        {} & {} & 0.70 &     11,734 \\
        {} & {} & 0.80 &      3,496 \\
    \midrule
        \multirow{3}{*}{\CLEFDataset} & Train & - & 999 \\
        {} & Dev  & - & 199 \\
        {} & Test  & - & 202 \\
    \bottomrule
    \end{tabular}
    }
    \caption{Statistics about our collected datasets in terms of {tweet--verifying article} pairs. 
    }
    \label{tab:datasplits}
\end{table}

\subsection{Baselines and State-of-the-Art}
\label{sec:appx:baselines}
\paragraph{Retrieval} Following \cite{DBLP:conf/clef/ShaarHMHBAMEN21}, we use an information retrieval model based on BM25~\citep{10.1561/1500000019} that ranks the fact-checking articles based on the relevance score between their \emph{\{'claim', 'title'\}} and the tweet.

\paragraph{Sentence-BERT} is a bi-encoder model based on Sentence-BERT
fine-tuned for detecting previously fact-checked claims using MNR loss. The details are in Section~\ref{sec:method:sbert}, \emph{General Scheme}.

\paragraph{Team DIPS}~\cite{DBLP:conf/clef/MihaylovaBCHHN21} adopts a Sentence-BERT model that computes the cosine similarity for each pair of an input tweet and a verified claim (article). The final ranking is made by passing a sorted list of cosine similarities to a fully-connected neural network.

\paragraph{Team NLytics}~\cite{Pritzkau2021NLyticsAC} uses a RoBERTa-based model optimized as a regression function obtaining a direct ranking for each tweet-article pair.

\paragraph{Team Aschern}~\cite{DBLP:conf/clef/ChernyavskiyIN21} combines TF.IDF with a Sentence-BERT (ensemble with three models of each type). The final ranking is obtained from a re-ranking LambdaMART model.

\subsection{Experimental Results}

Below, we present experiments that (\emph{i})~aim to analyze the impact of training with the distantly supervised data from \emph{\OurDataset}, and (\emph{ii})~to further improve the state-of-the-art (SOTA) results using modeling techniques to better leverage the noisy examples (see Section~\ref{sec:method}). In all our experiments, we evaluate the model on the development and on the testing sets from \CLEFDataset (see Table~\ref{tab:datasplits}), and we train on a mix with \emph{\OurDataset}. The reported results for each experiment (for each metric) are averaged over three runs using different seeds.

\begin{table}[t]
    \centering
    \setlength{\tabcolsep}{3pt}
    \resizebox{.5\textwidth}{!}{%
    \begin{tabular}{lrrr}
    \toprule
    \bf{Model} &  \bf{MRR}  & \bf{P@1} & \bf{MAP@5} \\
    \midrule
    \multicolumn{4}{c}{\bf{Baselines (\CLEFDataset)}} \\
    Retrieval~{\small{\citep{DBLP:conf/clef/ShaarHMHBAMEN21}}} & 76.1 & 70.3 & 74.9 \\
    SBERT (\CLEFDataset) & 79.96 &        74.59 &  79.20  \\
    
    \midrule
    \multicolumn{4}{c}{\bf{\OurDataset (Our Dataset)}} \\
                                SBERT (jac > 0.30) & 81.50 &        76.40 & 80.84 \\
                                SBERT (cos > 0.50) & 81.58 &        75.91 & 81.05 \\
    \midrule
    \multicolumn{4}{c}{\bf{(Pre-train) \OurDataset, (Fine-tune) \CLEFDataset}} \\
                  SBERT (jac > 0.30, Seq) & \bf{83.76} &        \bf{78.88} &  \bf{83.11}  \\
                  SBERT (cos > 0.50, Seq) & 82.26 &        77.06 & 81.41 \\
    
    \midrule
    \multicolumn{4}{c}{\bf{(Mix) \OurDataset and \CLEFDataset}} \\
             SBERT (jac > 0.30, Mix)  & 83.04 &        78.55 &    82.30 \\
             SBERT (cos > 0.50, Mix) & 82.12 &        76.57 &  81.38 \\
    \bottomrule
    \end{tabular}
    }
    \caption{Evaluation on the \CLEFDataset test set. In parenthesis is the name of the training split, i.e.,~\emph{Jac}card
    or \emph{Cos}ine 
    selection
    strategy, \emph{(Seq)} first training on \OurDataset and then on \CLEFDataset, \emph{(Mix)} mixing the data from the two. The best results are in \textbf{bold}.}
    \label{tab:sbert:distsup}
    \vspace{-0.2cm}
\end{table}

\paragraph{{Threshold Selection Analysis}}
Our goal here is to evaluate the impact of using distantly supervised data from \OurDataset. In particular, we fine-tune an SBERT baseline, as described in Section~\ref{sec:method:sbert}, using four different strategies: (\emph{i})~fine-tune on the training data from \CLEFDataset, (\emph{ii})~fine-tune on \emph{\OurDataset}, (\emph{iii})~pre-train on 
\emph{\OurDataset} and then fine-tune on the training data from \CLEFDataset, (\emph{iv})~mixing the data from both datasets. 

Table~\ref{tab:sbert:distsup} shows the results grouped based on training data used. In each group, we include the two best-performing models. 
We see that all SBERT models outperform the Retrieval baseline by 4--8 MAP@5 points absolute. Interestingly, training only on distantly supervised data is enough to outperform the SBERT model trained on the \CLEFDataset by more than 1.5 MAP@5 points absolute. 
Moreover, the performance of both data labeling strategies (i.e., Jaccard and Cosine) is close, suggesting comparable amount of noise in them. 

Next, we train on combined data from the two datasets. Unsurprisingly, both mixing the data and training on the two datasets sequentially 
{(\OurDataset $\xrightarrow{}$ \CLEFDataset)}
yields additional improvement compared to training on a single dataset. 
We achieve the best result when the model is first pre-trained on the \emph{({jac} > 0.3)} subset of \OurDataset, and then fine-tuned on \CLEFDataset: it improves by two points absolute in all measures
compared to \emph{SBERT (\OurDataset)}, and by four points compared to \emph{SBERT (\CLEFDataset)}. 

Nevertheless, we must note that pre-training with the \emph{Cosine similarly (cos > 0.50)} did not yield such sizable improvements as the ones when using Jaccard. We attribute this, on one hand, to the higher expected noise in the data according to our manual annotations (see Section~\ref{sec:feasability}), and on the other hand, to these examples being annotated by a similar model, and thus presumably easy for it.

\begin{figure}[t]
    \centering
    \includegraphics[width=0.95\columnwidth]{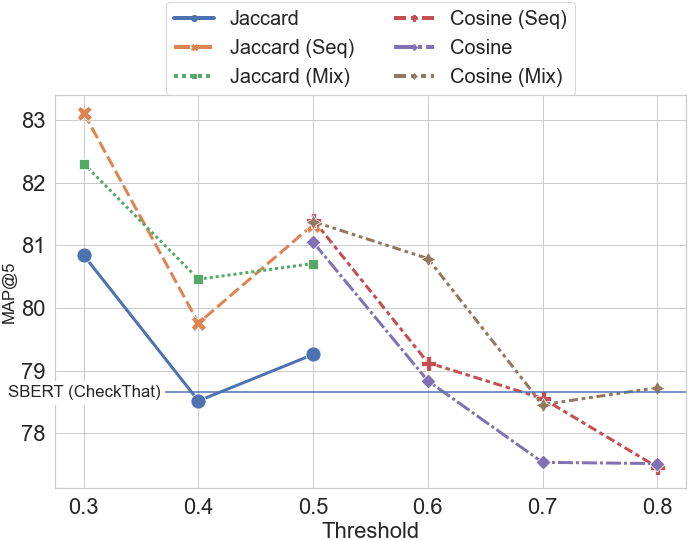}
    \caption{MAP@5 for different thresholds and distant supervision approaches. The \emph{Jaccard} and the \emph{Cosine} models are trained only on \OurDataset, while (\emph{Seq}) and (\emph{Mix}) were trained also on \CLEFDataset.
    }
    \label{fig:map5}
\end{figure}

We further analyze the impact of choosing different thresholds for the distant supervision approaches. Figure~\ref{fig:map5} shows the change of MAP@5 for each data labeling strategy. On the left, in the interval [0.3--0.5], are shown the results of the Jaccard-based
data labeling strategy, and on the right ([0.5--0.8]) are for the Cosine strategy. Once again, the models trained on the data selected using Jaccard similarity perform similarly or better than the \emph{SBERT (\CLEFDataset)} model (blue solid line). On the other hand, the Cosine-based selection outperforms the baseline only in small thresholds $\leq$ 0.6. These observations are in favor of the hypothesis that the highly ranked pairs from the fine-tuned SBERT model are easy examples, and do not bring much signal to the model over the \CLEFDataset data, whereas the Jaccard ranked ones significantly improve the model's performance. We further see similar performance when training with data from the lowest two thresholds for the two similarities (without data mixing), which suggests that these subsets have similar characteristics. 

Adding more distantly supervised data is beneficial for the model, regardless of the strategy. The only exception is the drop in performance when we decrease the Jaccard threshold from 0.5 to 0.4.

We attribute this to the quality of the data in that bracket, as the examples with lower similarity are expected to add more noise. However, the results improve drastically at the next threshold (which also doubles the number of examples), i.e.,~the model can generalize better from the new data. There is no such drop in the Cosine strategy. We explain this with expectation that noise increases proportionally to the decrease in model confidence. 

Finally, we report the performance of each model both on the development and on the test sets in Appendix~\ref{sec:appx:results}, Tables~\ref{tab:appx:dev_results} and~\ref{tab:appx:test_results}.

\begin{table}[t]
    \centering
    \setlength{\tabcolsep}{3pt}
    \resizebox{.5\textwidth}{!}{%
    \begin{tabular}{lccc}
        \toprule
        \bf{Model} & \multicolumn{2}{c}{\bf{MAP@5}} \\
        {}    &  Dev & Test \\
        \midrule
        DIPS~{\small{\citep{DBLP:conf/clef/MihaylovaBCHHN21}}} & {93.6} & 78.7 \\
        NLytics~{\small{\citep{Pritzkau2021NLyticsAC}}} & - & 79.9 \\
        Aschern~{\small{\citep{DBLP:conf/clef/ChernyavskiyIN21}}} & {94.2} & 88.2  \\ 
        \midrule
        SBERT (jac > 0.30, Mix) & 90.0 & 82.3 \\
        \hspace{5pt}+ shuffling \& trainable temp. & 92.4 & 82.6  \\
        \hspace{10pt}+ self-adaptive training (Eq.~\ref{eq:1}) & 92.6 & 83.6  \\
        \hspace{15pt}+ loss weights  & \textit{92.7} & \textit{84.3}  \\
        \midrule
        \hspace{20pt}+ TF.IDF + Re-ranking & 93.1 & 89.7 \\
        \hspace{25pt}+ TF.IDF + Re-ranking (ens.) & \bf 94.8 & \bf{90.3} \\
        \bottomrule
    \end{tabular}
    }
    \caption{Results on \CLEFDataset (dev and test). We compare our model and its components (added sequentially) to the state of the art. The best results are in \textbf{bold}.}
    \label{tab:noise}
\end{table}

\paragraph{Modeling Noisy Data}
We explore the impact of the proposed changes to the SBERT training approach: (\emph{i})~shuffling and training temperature, (\emph{ii})~data-related modification of the MNR loss for self-adaptive training with weights. We use the \textit{(jac > 0.30, mix)} approach in our experiments, as the baseline SBERT models achieved the highest scores on the dev set (Table~\ref{tab:appx:dev_results}). In Table~\ref{tab:noise}, we ablate each of these modifications by adding them iteratively to the baseline SBERT model. 

First, we can see that adding a special shuffling procedure and a trainable temperature ($\tau$) improves the MAP@5 by 2 points on the dev set and by 0.3 points on the test set.  
Next, we see a sizable improvement of 1 MAP@5 point on the test set, when using the self-adaptive training with MNR loss. Moreover, an additional 
0.7 points come from adding weights to the loss,
arriving at MAP@5 of 84.3. These weights allow the model to give higher importance to the less noisy data during training. 

Note that for these two ablations the improvements on the development set are diminishing. We attribute this to its small size (199 examples) and to the high values of MAP@5. Finally, note that our model without re-ranking outperforms almost all state-of-the-art models (except for that of team Aschern) by more than 4.5 points on the test dataset.

The last two rows of Table~\ref{tab:noise} show the results of our model that includes all proposed components, in combination with TF.IDF features and the LambdaMART re-ranking, described in Section~\ref{sec:method}.
Here, we must note that our model is trained on part of the \CLEFDataset training pool (80\%) -- the other part is used to train the re-ranking model. 
The full setup boosts the model's MAP@5 to \emph{89.7} when using a single model of the TF.IDF and SBERT (using the title/subtitle/claim as inputs, same as SBERT). With the ensemble architecture (re-ranking based on the scores of three TF.IDF and three SBERT models), we achieve our best results of \emph{90.3} on the test set (adding 1.7 MAP@5 on dev, and 0.6 on test), outperforming the previous state-of-the-art approach (\emph{Aschern},
88.2) by 2 MAP@5 points, and by more than 11 compared to the second best model~(\emph{NLytics},
79.9). This improvement corresponds to the observed gain over the SBERT model without re-ranking. Nevertheless, the change in the strength of the factors in LambdaMART is less. The TF-IDF models still have high importance for re-ranking -- a total of 41\% compared to 42.8\% reported in \citet{DBLP:conf/clef/ChernyavskiyIN21}. Here, we have a decrease mainly due to an increase of the importance of the reciprocal rank factor from 18.8\% to 20.2\% of the SBERT model that selects candidates.

\section{Discussion}
\label{sec:discussion}

Our proposed distant supervision data selection strategies show promising results, achieving SOTA results on the \CLEFDataset. Nonetheless, we are not able to identify all matching pairs in the list of candidates in \OurDataset. Hereby, we try to estimate their number using statistics from our manual annotations,\footnote{Due to the small number of annotated examples the variance in the estimates is large.} as shown in Tables~\ref{tab:wo_data_stats} and \ref{tab:cos_data_stats}. 

In particular, we estimate it by multiplying the fraction of {correct} pairs in each similarity bin by the number of examples in this bin. Based on cosine similarity, we estimate that out of the 332,600 pairs, the matching pairs are approximately 90,170 (27.11\%). 

Based on the Jaccard distribution, we estimate that 14.79\% of all tweet-conversations (root of the conversation), and 22.23\% of the tweet--reply (the tweet before the current one in the conversation) pairs are good, or nearly 61,500 examples.

Our experiments show that the models can effectively account for the noise in the training data. The self-adaptive training and the additional weighing in the loss (described in Section~\ref{sec:method}) yield 1 additional MAP@5 point each. This suggests that learning from noisy labels~\cite{10.5555/3327757.3327944,wang-etal-2019-learning-noisy,song2020learning,zhou-chen-2021-learning} and using all examples in \OurDataset can improve the
results even further. Moreover, incorporating the negative examples (non-matching pairs) from \OurDataset in the training could also help~\cite{lu-etal-2021-multi,thakur-etal-2021-augmented}.

\section{{Related Work}}
\label{sec:relatedwork}

\paragraph{Previously Fact-Checked Claims}
While fake news and mis/disinformation detection have been studied extensively~\cite{10.1145/2897350.2897352,10.1145/3161603,ijcai2020-672,COLING2022:Multimodal,guo-etal-2022-survey,hardalov-etal-2022-survey}, the problem of detecting previously fact-checked claims remains under-explored. \citet{10.14778/3137765.3137815} mentioned the task as a component of an end-to-end fact-checking pipeline, but did not evaluate it nor studied its contribution.
\citet{hossain-etal-2020-covidlies} retrieved evidence from a list of known misconceptions and evaluated the claim's veracity based on its stance towards the hits; while this task is similar, it is not about whether a given claim was fact-checked or not.

Recently, the task received more attention. 
\citet{shaar-etal-2020-known} collected two datasets, from PolitiFact (political debates) and Snopes (tweets), of claims and corresponding fact-checking articles. 
The CLEF \emph{CheckThat!} lab~\citep{clef-checkthat:2020,barron2020overview,CheckThat:ECIR2020,CheckThat:ECIR2021,clef-checkthat:2021:LNCS,DBLP:conf/clef/ShaarHMHBAMEN21,ECIR:CLEF:2022,clef-checkthat:2022:LNCS,clef-checkthat:2022:task2} extended these datasets with more data in English and Arabic. 
The best systems \citep{Pritzkau2021NLyticsAC,DBLP:conf/clef/MihaylovaBCHHN21,chernyavskiy-etal-2022-batch} used a combination of BM25 retrieval, semantic similarity using embeddings~\cite{reimers-gurevych-2019-sentence}, and reranking. \citet{bouziane2020team} used extra data from fact-checking datasets~\cite{wang-2017-liar,thorne-etal-2018-fever,wadden-etal-2020-fact}. 

Finally, \citet{Claim:retrieval:context:2022} and \citet{EMNLP2022:All:Claims:Document} explored the role of the context in detecting previously fact-checked claims in political debates.

Our work is most similar to that of \citet{vo-lee-2020-facts}, who mined 19K tweets and corresponding fact-checked articles. Unlike them, we focus on textual claims (they were interested in multimodal tweets with images), we collect an order of magnitude more examples, and we propose a novel approach to learn from such noisy data directly (while they manually checked each example).

\paragraph{Training with Noisy Data} {Leveraging} large collections of unlabeled data has been at the core of large-scale language models using Transformers~\citep{nips2017_7181:transformer}, 
such as GPT~\cite{radford2018gpt1,radford2019language}, BERT~\cite{devlin2019bert}, and RoBERTa~\cite{liu2019roberta}. Recently, such models used noisy retrieved data~\cite{NEURIPS2020_6b493230,pmlr-v119-guu20a} or active relabeling and data augmentation~\cite{thakur-etal-2021-augmented}.
Distant supervision is also a crucial part of recent breakthroughs in few-shot learning~\cite{schick-schutze-2021-exploiting,schick-schutze-2021-just}.

Yet, there has been little work of using noisy data for fact-checking tasks. \citet{10.1145/3331184.3331248} collected tweets containing a link to a fact-checking website, based on which they tried to learn a fact-checking language and to generate automatic answers. \citet{you2019attributed} used similar data from tweets for fact-checking URL recommendations. 

Unlike the above work, here we propose an automatic procedure for labeling and self-training specifically designed for the task of detecting previously fact-checked claims.

\section{Conclusion and Future Work}
\label{sec:conclusion}

We presented \OurDataset, a large dataset for detecting previously fact-checked claims, with more than 330,000 pairs of tweets and corresponding fact-checking articles posted by crowd fact-checkers. We further investigated two techniques for labeling the data using distance supervision,
resulting in training sets of 3.5K--50K examples. We also proposed an approach for training from noisy data using self-adaptive learning and additional weights in the loss function. Furthermore, we demonstrated that our data yields sizable performance gains of four points in terms MRR, P@1, and MAP@5 over strong baselines.
Finally, we demonstrated improvements over the state of the art on the \CLEFDataset test set
by two points, 
when using our proposed dataset and pipeline.

In future work, we plan to experiment with more languages and more distant supervision techniques such as predictions from an ensemble model.

\section*{Acknowledgments}

We want to thank Ivan Bozhilov, Martin Vrachev, and Lilyana Videva for the useful discussions and for their help with additional data analysis and manual annotations. 

This work is partially supported by Project UNITe BG05M2OP001-1.001-0004 funded by the Bulgarian OP ``Science and Education for Smart Growth'', co-funded by the EU via the ESI Funds.

The work was prepared within the framework of the HSE University Basic Research Program.

\section*{Ethics and Broader Impact}

\subsection*{Dataset Collection}

We collected the dataset using the Twitter API.\footnote{We use the Twitter API v2 with \href{https://developer.twitter.com/en/products/twitter-api/academic-research}{academic research access}, \url{http://developer.twitter.com/en/docs},}
following the terms of use outlined by Twitter.\footnote{\url{http://developer.twitter.com/en/developer-terms/agreement-and-policy}} Specifically, we only downloaded public tweets, and we only distribute dehydrated Twitter IDs. 

\subsection*{Biases}

We note that some of the annotations are subjective, and we have clearly indicated in the text which these are. Thus, it is inevitable that there would be biases in our dataset. Yet, we have a very clear annotation schema and instructions, which should reduce the biases.

\subsection*{Misuse Potential}

Most datasets compiled from social media present some risk of misuse. We, therefore, ask researchers to be aware that our dataset can be maliciously used to unfairly moderate text (e.g.,~a tweet) that may not be malicious based on biases that may or may not be related to demographics and other information within the text. Intervention with human moderation would be required in order to ensure this does not occur.

\subsection*{Intended Use}

Our dataset can enable automatic systems for analysis of social media content, which could be of interest to practitioners, professional fact-checker, journalists, social media platforms, and policymakers. Such systems can be used to alleviate the burden of moderators, but human supervision would be required for more intricate cases and in order to ensure that no harm is caused.

Our models can help fight the COVID-19 infodemic, and they could support analysis and decision-making for the public good. However, the models can also be misused by malicious actors. Therefore, we ask the users to be aware of potential misuse. With the possible ramifications of a highly subjective dataset, we distribute it for research purposes only, without a license for commercial use. Any biases found in the dataset are unintentional, and we do not intend to do harm to any group or individual.

\subsection*{Environmental Impact}
We would like to warn that the use of large-scale Transformers requires a lot of computations and the use of GPUs/TPUs for training, which contributes to global warming \cite{strubell-etal-2019-energy}. This is a bit less of an issue in our case, as we do not train such models from scratch; rather, we fine-tune them on relatively small datasets. Moreover, running on a CPU for inference, once the model has been fine-tuned, is perfectly feasible, and CPUs contribute much less to global warming.

\bibliography{bibliography}

\appendix

\section{Hyperparameters and Fine-Tuning}
\label{sec:appx:hyper:ft}

Below, we first describe the common parameters we use, and then we give the values of model-specific parameters.

\paragraph{Common Parameters}
\begin{itemize}%
    \item We develop our models in Python using PyTorch~\citep{NEURIPS2019_9015}, the Transformers library~\citep{wolf-etal-2020-transformers}, and the Sentence Transformers library.~\citep{reimers-gurevych-2019-sentence}\footnote{\url{http://github.com/UKPLab/sentence-transformers}}
    \item We used NLTK~\citep{loper-2002-nltk} to filter out English stop words, the \emph{Twitter Tokenizer} to split the tweets and to strip the handles, and the Porter stemmer~\citep{porter1980algorithm} to stem the tokens.
    \item For optimization, we use AdamW~\citep{loshchilov2018decoupled} with weight decay of 1e-8, $\beta_1$ = 0.9, $\beta_2$ = 0.999, $\epsilon$ = 1e-08, for 10 epochs, and maximum sequence length of 128 tokens (per encoder).\footnote{When needed, we truncated the sequences token by token, starting from the longest sequence in the pair.}
    \item All Sentence BERT (SBERT) models are initialized from the \texttt{stsb-bert-base}\footnote{\url{huggingface.co/sentence-transformers/stsb-bert-base}} checkpoint.
    \item The SBERT models use cosine similarity both during training inside the MNR loss and during inference for ranking.
    \item We selectd the values of the hyper-parameters on the development set of \CLEFDataset,\footnote{\url{https://gitlab.com/checkthat_lab/clef2021-checkthat-lab/-/tree/master/task2}} and we chose the best model checkpoint based on the performance on the development set (MAP@5).
    \item We ran each experiment three times with different seeds and averaged the result scores.
    \item The models were evaluated on each epoch or every 250 steps, whichever is less.
    \item The evaluation measures are calculated using the official code from the \CLEFDataset competition~\cite{DBLP:conf/clef/ShaarHMHBAMEN21}\footnote{\url{https://gitlab.com/checkthat_lab/clef2021-checkthat-lab/-/tree/master/task2/scorer}} and the SentenceTransformer's library.
    \item In our work, we list 199 examples for the development set of \CLEFDataset, while \citet{DBLP:conf/clef/ShaarHMHBAMEN21} lists 200. The difference comes from one duplicate row in the development set, which we found and filtered out.
    \item We trained our models on 5x Tesla T4 GPUs and 1x GeForce GTX 1080Ti, depending on the dataset size, the experiments took between 10 minutes and 5 hours.
\end{itemize}

\paragraph{Baseline SBERT}
\begin{itemize}%
    \item Our baseline Sentence BERT is trained with LR of 2e-05, warmup of 0.1, and batch size of 32.
    \item We set the temperature ($\tau$) in the MNR loss to 1.0, i.e.,~using unmodified MNR.
    \item The model consists of 110M params, same as the bert-base~\citet{devlin2019bert}, as it uses a bi-encoder scheme.
\end{itemize}

\paragraph{Proposed Pipeline}
\begin{itemize}%
    \item The model is trained with LR of 1e-05, warmup of 0.1, batch size of 8, ad group size of 4 during the dataset shuffling.
    \item We tuned the settings of the self-adaptive training, and ended up with the folowing values: momentum $\alpha$ of 0.9, refurbishment process starting at the second epoch.
    \item We set the learning rate for the temperature ($\tau$) in the MNR loss to 0.4.
    \item In the re-ranking, we used 800 training examples to train SBERT and the remaining 199 examples to train LambdaMART.
    \item We re-ranked the top-100 results from the best SBERT model with LambdaMART. 
    \item All other training details we kept from \cite{DBLP:conf/clef/ChernyavskiyIN21}.
    \item The model has 330M params, 3x as the size of the Baseline SBERT, as it trains three separate models.
    \item In our preliminary experiments, SBERT-base and SBERT-large yielded the same results in terms of MAP@5, ad thus we experiment with the \textit{base} versions.
\end{itemize}

\section{Dataset}

Below, we first give some detail about the process of article collection, and then we discuss the overlap between our \OurDataset dataset and \CLEFDataset.

\subsection{Fact-checking Articles Collection} 
\label{sec:appx:factcheckcollect}

In order to obtain a collection of fact-checking articles for each tweet, we first formed a list of unique URLs shared in the fact-checking tweets from the crowd fact-checkers. Next, from each URL we downloaded the HTML of the whole page and extracted the meta information using CSS selectors and RegEx rules. In particular, we followed previous work~\cite{barron2020overview,DBLP:conf/clef/ShaarHMHBAMEN21} and collected: \emph{title} (the title of the page), \emph{subtitle} (short description of the fact-check), \emph{claim} (the claim of interest), \emph{subtitle} (short description of the fact-check), \emph{date} (the date the article was published), and \emph{author} (the author of the article). We do not parse the content of the article and the factual label, as the credibility of the claim is not related to the objective of this task, i.e.,~the goal is to find a fact-checking article, but not to verify it.

As a result, we collected 10,340 articles that were published in the period between 1995--2021. The per-year distribution is shown in Table~\ref{fig:snopes_dist} (in brown). The majority of the articles are from the period after 2015, with a peak at the ones from 2020/2021. We attribute this on the increased media literacy and on the nature of the Twitter dynamics~\cite{zubiaga2018longitudinal}.

\subsection{\CLEFDataset Word Overlaps}
\label{sec:appx:dataanalysis}

\begin{figure}[t]
    \centering
    \includegraphics[width=\columnwidth]{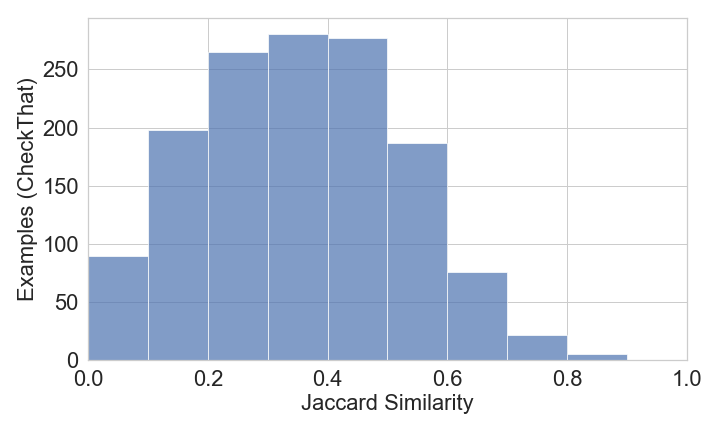}
    \caption{Distribution of the Jaccard similarity scores. The score is an average of the \emph{sim(tweet, title)} and \emph{sim(tweet, subtitle)}.}
    \label{fig:clef_wo}
\end{figure}

Next, we analyzed the distribution of the Jaccard scores in the \CLEFDataset, shown in Figure~\ref{fig:clef_wo}. The distribution is different compared to the one observed in our newly collected dataset, as it peaks at around 0.4, and is slightly shifted towards lower similarity values, suggesting that the examples included are not easily solvable with basic lexical features~\cite{DBLP:conf/clef/ShaarHMHBAMEN21}, which we also observe in our experiments (see Section~\ref{sec:experiments}).

\section{Annotations}
\label{sec:appx:annotations}

\paragraph{Setup and Guidelines}
Each annotator was provided with the guidelines and briefed by one of the authors of this paper. For annotation, we used a Google Sheets document, where none of the annotators had access to the annotations by the others.

The annotation sheet contained the following fields:
\begin{itemize}%
    \item \textit{tweet\_text}: the text of the fact-checking tweet;
    \item \textit{text\_conversation}: the text of the root of the conversation;
    \item \textit{text\_reply}: the text of the last tweet before the fact-checking one;
    \item \textit{title}: the title of the Snopes article;
    \item \textit{subtitle}: the subtitle of the Snopes article.
\end{itemize}

The annotation task was to mark whether the \emph{conversation matches} and also whether the \emph{reply matches} using check-boxes. We also allowed the annotators to add comments as a free-form text. 

\paragraph{Demographics} 
We recruited three annotators: two male and one female, between 25 and 30 years old, with higher education (at least a bachelors degree), and currently enrolled in a MSc or PhD programs in Computer Science. Each annotator was proficient in English, but they were not native speakers.

\paragraph{Inter-Annotator Agreement}
Here, we present the inter-annotator agreement. We measure the overall agreement using Fleiss kappa~\citep{fleiss1971measuring} (shown in Figure~\ref{tab:fleiss}) and also the agreement between each two annotators using Cohen's Kappa (shown in Table~\ref{tab:cohen}). The overall level of agreement between the annotators is \emph{good}. Moreover, we can see that between annotator A and C the agreement is almost perfect both for the replies and for the conversations. The lowest agreement is between A and B, but it is still substantial.

\begin{table}[tbh]
    \centering
    \begin{tabular}{lrr}
    \toprule
     & \bf  Replay  &  \bf Conversation   \\
    \midrule
    Fleiss Kappa &   0.745 & 0.750 \\
    \bottomrule
    \end{tabular}
    \caption{Fleiss Kappa inter-annotator agreement between all three of our annotators: A, B, and C.}
    \label{tab:fleiss}
\end{table}

\begin{table}[tbh]
    \centering
    \begin{tabular}{lrr}
    \toprule
                \bf Annotators & \bf Replay  &  \bf Conversation    \\
                {} & \multicolumn{2}{c}{\bf Cohen's Kappa}\\
    \midrule
    A $\leftrightarrow$ B &   0.650 & 0.655 \\
    A $\leftrightarrow$ C &   0.885 & 0.922 \\
    B $\leftrightarrow$ C &   0.698 & 0.673 \\
    \bottomrule
    \end{tabular}
    \caption{Cohen's Kappa pairwise inter-annotator agreement between all pairs of our annotators.}
    \label{tab:cohen}
\end{table}

\paragraph{Disagreement Analysis}
After the annotations procedure was finished, we analyzed the examples for which the annotators disagreed, which fell in the following categories:

\begin{enumerate}[label={(\roman*)}]
    \item Claims depending on information from external sources, e.g., \emph{`Blame Russia again? [URL]'}.
    \item Tweets containing multiple claims, for which the referenced article does not target the main claim, e.g., \emph{```It sounds like someone who is scared as heck that they will not win,'' Shermichael Singleton says of Pres. Trump’s remarks encouraging his supporters to vote twice.'} Here, the corresponding crowd fact-check is \emph{`Did Trump Tell People To Vote Twice?'}, i.e.,~the main claim is in the quote itself, while the remark about voting twice is secondary.
    \item The claim is ambiguous, e.g.,~\emph{`Fanta (soft drink) was created so that the Nazi's could replace Coca-Cola during WWII [URL]'}, and the fact-check is about \emph{`Was Fanta invented by the Nazis?'}. Here, it is not clear who created Fanta.
    \item The claim is a partial match, e.g.,~\emph{`did President Trump have a great economy and job creation for 1st 3 years???'}, and the fact-check is \emph{`Did Obama’s Last 3 Years See More Jobs Created Than Trump’s First 3?'}, which only covers part of the claim in the tweet.
\end{enumerate}

\begin{table*}[t!]
    \centering
    \resizebox{\textwidth}{!}{%
    \begin{tabular}{p{0.02\columnwidth}p{0.9\columnwidth}p{0.9\columnwidth}}
         \toprule
         {} &{\bf Tweet w/ Claim} & {\bf Snopes Verified Claim and Article} \\
         \midrule
         \multicolumn{3}{c}{\bf Correct Matches \correctmark} \\
         (1) & “Mussolini may have done many brutal and tyrannical things; he may have destroyed human freedom in Italy; he may have murdered and tortured citizens whose only crime was to oppose Mussolini; but ‘one had to admit’ one thing about the Dictator: he ‘made the trains run on time.’” [URL] &  Italian dictator Benito Mussolini made the trains run on time \url{snopes.com/fact-check/loco-motive/} \\
         \hline
         (2) & "Full list of songs Clear Channel banned following the 9\/11 attacks. 
         Some of these don’t make any sense at all. 1\/2 [URL]"& 
         Clear Channel Communications banned their American radio stations from playing specified songs in order to avoid offending listeners. \url{snopes.com/fact-check/radio-radio/} \\
         \hline
         (3) & @user @user OMG! Were you on this planet when Obama did nothing during H1N1 crisis? Only difference was H1N1 caused more than 10000 deaths and Obama was golfing. Took 6 mos for him to even have a press conference! &  U.S. President Barack Obama waited until millions were infected and thousands were dead before declaring a public health emergency concerning swine flu. \url{snopes.com/fact-check/obama-wait-swine-flu-n1h1/} \\
         \midrule
         \multicolumn{3}{c}{\bf Incorrect Matches \incorrectmark} \\
         (4) & Dick Van Dyke?  What's next?  Penis Van Lesbian?  What.  Is.  NEXT??? &  Dick Van Dyke’s real name is Penis Van Lesbian. \url{snopes.com/fact-check/dick-van-dyke/} \\
         \hline
         (5) & "I've just found a 2012 report on how well informed TV viewers are NPR was top, of course. That's the one the Republicans want to defund, as it's contrary to their interests Also Fox viewers were less well informed than people who did not watch TV news at all" & A four-year study has found that Fox News viewers have IQs 20 points lower than average. \url{snopes.com/fact-check/news-of-the-weak/} \\
         \hline
         (6) & Trump just said he has seen gas prices at \$.89-\$.99 per gallon. Where I am it is currently \$1.70. Anyone see prices Trump is talking about? & The COVID-19 coronavirus disease is "spreading quickly from gas pumps." \url{snopes.com/fact-check/covid19-gas-pump-handles/} \\
         \bottomrule
    \end{tabular}
    }
    \caption{{Examples from \OurDataset, showing correct ({\correctmark}) and incorrect matches ({\incorrectmark}). The examples in each group are sorted by their overlap with the claim made in the tweet.}}
    \label{tab:annotation_examples}
\end{table*}

\paragraph{{Tweet-Article Pairs Analysis}}
In Table~\ref{tab:annotation_examples}, we show examples of \emph{correct} ({\correctmark}) and \emph{incorrect} ({\incorrectmark}) matching pairs. We sorted the examples within each group based on the word overlap between the claim and the verified claim, e.g.,~(1) and (2) have more words in common between the two texts compared to the overlaps in (3), and similarly for (4)--(6).

First, we can see that high overlap does not guarantee a correct matching tweet--article pair, just like low overlap does not mean an incorrect pair, which is also visible from the analysis of the Jaccard similarity in Table~\ref{tab:wo_data_stats}. These two phenomena can be seen in (\emph{3}), which contains a correct pair with low overlap, and in (\emph{4}), where there is an incorrect match with high overlap. Next, some tweets may not contain a claim such as (\emph{4}), as the user only asks questions, rather than stating something that can be fact-checked. In contrast, (\emph{6}) contains a verifiable claim about \emph{gas prices}, but the linked Snopes article fact-checks whether \emph{COVID spreads through gas pumps}, which is irrelevant in this case. Row (\emph{5}) is a partial match, and the tweet contains a check-worthy claim, but the article by the crowd fact-checker focuses on the IQ of the Fox News viewers, rather than on how well informed they are, and thus again the match is incorrect. Finally, in row (\emph{1}), we can see that the verified claim is almost exactly included in the tweet, which is an easy case to match. In contrast, for the example in row (\emph{3}), the model should do a semantic match based on some prior knowledge that the other name for \emph{influenza A virus subtype H1N1} is \emph{swine flu}, and moreover, \emph{10,000} should be associated with the word \emph{thousands}.

\section{Experimental Results}

\begin{table*}[t]
    \centering
    \setlength{\tabcolsep}{3pt}
    \begin{tabular}{lrrr}
    \toprule
    \bf{Model} &  \bf{MRR}  & \bf{P@1} & \bf{MAP@5} \\
    \midrule
    \multicolumn{4}{c}{\bf{Baselines (\CLEFDataset)}} \\
    Retrieval~{{\citep{DBLP:conf/clef/ShaarHMHBAMEN21}}} & 76.1 & 70.3 74.9  \\
    SBERT (\CLEFDataset) & 87.97 &        84.92 & 87.45  \\
    \midrule
    \\
    \multicolumn{4}{c}{\bf{\OurDataset (Our Dataset)}} \\
    \midrule
    SBERT (cos > 0.50) & 88.20 &        85.76 &   87.80 \\
    SBERT (cos > 0.60) & 87.21 &        84.25 &   86.69 \\
    SBERT (cos > 0.70) & 86.18 &        83.08 &   85.76 \\
    SBERT (cos > 0.80) & 83.57 &        80.40 &   82.93 \\
    SBERT (jac > 0.30) & 88.01 &        85.09 &   87.61 \\
    SBERT (jac > 0.40) & 87.26 &        84.76 &   86.80 \\
    SBERT (jac > 0.50) & 86.53 &        83.42 &   86.13 \\
    \midrule
    \\
    \multicolumn{4}{c}{\bf{(Pre-train) \OurDataset, (Fine-tune) \CLEFDataset}} \\
    \midrule
    SBERT (cos > 0.50, Seq) & 89.92 &        87.60 &   89.49 \\
    SBERT (cos > 0.60, Seq) & 89.56 &        87.27 &   89.20 \\
    SBERT (cos > 0.70, Seq) & 88.70 &        85.59 &   88.36 \\
    SBERT (cos > 0.80, Seq) & 88.42 &        85.26 &   88.03 \\
    SBERT (jac > 0.30, Seq) & 90.21 &        87.44 &   89.69 \\
    SBERT (jac > 0.40, Seq) & 89.64 &        86.77 &   89.25 \\
    SBERT (jac > 0.50, Seq) & 89.44 &        86.26 &   89.03 \\
    \midrule
    \\
    \multicolumn{4}{c}{\bf{(Mix) \OurDataset and \CLEFDataset}} \\
    \midrule
    SBERT (cos > 0.50, Mix) & 89.47 &        86.77 &   88.99 \\
    SBERT (cos > 0.60, Mix) & 88.54 &        85.76 &   87.98 \\
    SBERT (cos > 0.70, Mix) & 87.71 &        84.92 &   87.18 \\
    SBERT (cos > 0.80, Mix) & 88.40 &        85.26 &   87.97 \\
    SBERT (jac > 0.30, Mix) & 90.41 &        87.94 &   90.00 \\
    SBERT (jac > 0.40, Mix) & 89.82 &        86.60 &   89.48 \\
    SBERT (jac > 0.50, Mix) & 88.71 &        85.26 &   88.31 \\
    \bottomrule
    \end{tabular}
    \caption{Evaluation on the \CLEFDataset \textbf{development} set. In parenthesis is shown the name of the training split, i.e.,~Jaccard (\emph{jac}) or Cosine (\emph{cos}) for data selection strategy, \emph{(Seq)} for first training on \OurDataset and then on \CLEFDataset, and \emph{(Mix)} for mixing the data from the two datasets.}
    \label{tab:appx:dev_results}
\end{table*}

\label{sec:appx:results}

Here, we present the expanded results for our experiments described in Section~\ref{sec:experiments}. Tables~\ref{tab:appx:dev_results} and~\ref{tab:appx:test_results} include the results for the \emph{threshold selection analysis} experiments on the development dataset, and on the testing dataset, respectively. Here, Table~\ref{tab:appx:test_results} corresponds to Table~\ref{tab:sbert:distsup} in the main text of the paper, and includes all metrics and all thresholds (shown in Figure~\ref{fig:map5}). Next, the results from our \emph{Modeling Noisy Data} experiments are in Table~\ref{tab:sota:clef}, which corresponds to Table~\ref{tab:noise} in the main paper. In all tables, we use the same notation and grouping as in the corresponding table in the main paper.

\begin{table*}[t]
    \centering
    \setlength{\tabcolsep}{4pt}
    \resizebox{1.0\textwidth}{!}{%
    \begin{tabular}{lrrrrrrrrrrr}
    \toprule
    {} & {} & \multicolumn{5}{c}{\bf Precision} & \multicolumn{5}{c}{\bf MAP} \\
                             \bf Model &  \bf MRR  & \bf  @1 &  \bf @3 & \bf  @5 &  \bf @10 & \bf  @20 &  \bf @1 & \bf  @3 & \bf  @5 & \bf  @10 & \bf  @20 \\
    \midrule
    \multicolumn{12}{c}{\bf{Baselines (\CLEFDataset)}} \\
    Retrieval~{\small{\citep{DBLP:conf/clef/ShaarHMHBAMEN21}}} & 76.1 & 70.3 & 26.2 & 16.4 & 8.8 & 4.6 & 70.3 & 74.1 & 74.9 & 75.7 & 75.9  \\
    SBERT (\CLEFDataset) & 79.96 &        74.59 &        27.89 &        17.19 &          8.96 &          4.61 &  74.59 &  78.66 &  79.20 &   79.66 &   79.83 \\
                                                 
    \midrule
    \multicolumn{12}{c}{\bf{\OurDataset (Our Dataset)}} \\
    SBERT (cos > 0.50) & 81.58 &        75.91 &        28.60 &        17.76 &          9.04 &          4.67 &  75.91 &  80.36 &  81.05 &   81.27 &   81.48 \\
    SBERT (cos > 0.60) & 79.71 &        74.75 &        27.39 &        16.96 &          8.86 &          4.59 &  74.75 &  78.25 &  78.84 &   79.38 &   79.61 \\
    SBERT (cos > 0.70) & 78.27 &        72.28 &        27.61 &        17.10 &          8.89 &          4.53 &  72.28 &  76.95 &  77.54 &   78.01 &   78.12 \\
    SBERT (cos > 0.80) & 78.39 &        72.94 &        27.34 &        16.83 &          8.81 &          4.55 &  72.94 &  77.04 &  77.52 &   78.08 &   78.28 \\
    SBERT (jac > 30) & 81.50 &        76.40 &        28.49 &        17.43 &          8.94 &          4.65 &  76.40 &  80.45 &  80.84 &   81.14 &   81.38 \\
    SBERT (jac > 40) & 79.45 &        74.42 &        27.34 &        16.93 &          8.89 &          4.65 &  74.42 &  77.92 &  78.52 &   79.08 &   79.33 \\
    SBERT (jac > 50) & 79.96 &        74.75 &        27.89 &        17.29 &          8.94 &          4.60 &  74.75 &  78.63 &  79.26 &   79.63 &   79.81 \\
    \midrule
\multicolumn{12}{c}{\bf{(Pre-train) \OurDataset, (Fine-tune) \CLEFDataset}} \\
    SBERT (cos > 0.50, Seq)  & 82.26 &        77.06 &        28.27 &        17.62 &          9.26 &          4.76 &  77.06 &  80.64 &  81.41 &   81.99 &   82.18 \\
    SBERT (cos > 0.60, Seq)  & 80.13 &        75.41 &        27.45 &        17.00 &          8.94 &          4.65 &  75.41 &  78.55 &  79.13 &   79.76 &   79.99 \\
    SBERT (cos > 0.70, Seq)  & 79.27 &        73.43 &        27.72 &        17.33 &          8.94 &          4.58 &  73.43 &  77.78 &  78.56 &   78.94 &   79.09 \\
    SBERT (cos > 0.80, Seq)  & 78.32 &        72.77 &        27.17 &        16.93 &          8.89 &          4.58 &  72.77 &  76.71 &  77.41 &   77.98 &   78.15 \\
    SBERT (jac > 0.30, Seq)  & 83.76 &        78.88 &        28.93 &        17.82 &          9.21 &          4.71 &  78.88 &  82.59 &  83.11 &   83.49 &   83.63 \\
    SBERT (jac > 0.40, Seq)  & 80.69 &        75.25 &        27.83 &        17.33 &          9.09 &          4.69 &  75.25 &  79.04 &  79.76 &   80.34 &   80.57 \\
    SBERT (jac > 0.50, Seq)  & 81.99 &        76.90 &        28.16 &        17.76 &          9.13 &          4.69 &  76.90 &  80.34 &  81.33 &   81.70 &   81.88 \\
    \midrule
\multicolumn{12}{c}{\bf{(Mix) \OurDataset and \CLEFDataset}} \\
    SBERT (cos > 0.50, Mix) & 82.12 &        76.57 &        28.55 &        17.59 &          9.13 &          4.68 &  76.57 &  80.86 &  81.38 &   81.82 &   82.00 \\
    SBERT (cos > 0.60, Mix) & 81.45 &        76.40 &        28.27 &        17.43 &          8.96 &          4.61 &  76.40 &  80.25 &  80.79 &   81.14 &   81.31 \\
    SBERT (cos > 0.70, Mix) & 79.08 &        73.10 &        27.83 &        17.33 &          8.89 &          4.57 &  73.10 &  77.72 &  78.46 &   78.77 &   78.95 \\
    SBERT (cos > 0.80, Mix)  & 79.73 &        74.75 &        27.56 &        17.00 &          9.06 &          4.62 &  74.75 &  78.22 &  78.73 &   79.46 &   79.59 \\
    SBERT (jac > 0.30, Mix)  & 83.04 &        78.55 &        28.66 &        17.52 &          9.11 &          4.69 &  78.55 &  81.93 &  82.30 &   82.75 &   82.94 \\
    SBERT (jac > 0.40, Mix)  & 81.18 &        74.59 &        28.55 &        17.72 &          9.14 &          4.74 &  74.59 &  79.79 &  80.46 &   80.85 &   81.10 \\
    SBERT (jac > 0.50, Mix)  & 81.56 &        76.73 &        28.22 &        17.36 &          9.03 &          4.71 &  76.73 &  80.23 &  80.71 &   81.19 &   81.45 \\
    \bottomrule
    \end{tabular}
    }
    \caption{Evaluation on the \CLEFDataset \textbf{test} dataset. In parenthesis is shown the name of the training split: Jaccard (\emph{jac}) or Cosine (\emph{cos}) for data selection strategy, \emph{(Seq)} for first training on \OurDataset and then on \CLEFDataset, and \emph{(Mix)} for mixing the data from the two datasets.}
    \label{tab:appx:test_results}
\end{table*}

\begin{table*}[t]
    \centering
    \setlength{\tabcolsep}{4pt}
    \resizebox{1.0\textwidth}{!}{%
    \begin{tabular}{lrrrrrrrrr}
    \toprule
    {} & {} & \multicolumn{4}{c}{\bf Precision} & \multicolumn{4}{c}{\bf MAP} \\
                                                \bf  Model &  \bf MRR  & \bf  @1 & \bf  @3 & \bf  @5 & \bf  @10 & \bf  @1 & \bf  @3 & \bf  @5 & \bf @10 \\
    \midrule
    
    DIPS~{\small{\citep{DBLP:conf/clef/MihaylovaBCHHN21}}} & 79.5 & 72.8 & 28.2 & 17.7 & 9.2 & 72.8 & 77.8 & 78.7 & 79.1   \\
    NLytics~{\small{\citep{Pritzkau2021NLyticsAC}}} & 80.7 & 73.8 & 28.9 & 17.9 & 9.3  & 73.8 & 79.2 & 79.9 & 80.4   \\
    Aschern~{\small{\citep{DBLP:conf/clef/ChernyavskiyIN21}}} & 88.4 & 86.1 & 30.0 & 18.2 & 9.2  & 86.1 & 88.0 & 88.3 & 88.4 \\
    \midrule
    SBERT (jac > 0.30, Mix) & 83.0 &        78.6 &        28.7 &        17.5 &          9.1 &          78.6 &  81.9 &  82.3 &   82.8\\
\hspace{5pt}+ shuffling \& trainable temp. & 83.2 & 77.7 & 29.1 & 17.8 & 9.1 & 77.7 & 82.2 & 82.6 & 82.9  \\
\hspace{10pt}+ self-adaptive training (Eq.~\ref{eq:1}) & 84.2 & 78.7 & 29.3 & 18.1 & 9.3 & 78.7 & 83.0 & 83.6 & 83.9 \\
\hspace{15pt}+ loss weights  & 84.8 & 79.7 & 29.5 & 18.2 & 9.3 & 79.7 & 83.7 & 84.3 & 84.6  \\
\midrule
\hspace{20pt}+ TF.IDF + Re-ranking & 89.9 & 86.1 & 30.9 & 18.9 & 9.6 &  86.1 & 89.2 &    89.7 & 89.8  \\
\hspace{25pt}+ TF.IDF + Re-ranking (ens.) &  90.6 & 87.6 & 30.7 & 18.8 & 9.5 &  87.6 & 89.9 &    90.3 & 90.4  \\
    \bottomrule
    \end{tabular}
    }
    \caption{Results on the \CLEFDataset \textbf{test} dataset. We compare our model and its components (added sequentially) to three state-of-the-art approaches.}
    \label{tab:sota:clef}
\end{table*}

\end{document}